\documentclass[11pt,a4paper]{article}
\usepackage[hyperref]{acl2021}
\usepackage{times}
\usepackage{latexsym}

\usepackage{microtype}

\aclfinalcopy 
\def\aclpaperid{1639} 

\usepackage{url}            
\usepackage{amsfonts}       
\usepackage{nicefrac}       
\usepackage{booktabs}       
\usepackage{adjustbox}
\usepackage{graphicx}
\usepackage{color}
\usepackage{comment}
\usepackage{amsmath}
\usepackage{multirow}

\title{Cross-lingual Text Classification with Heterogeneous Graph Neural Network}

\author{Ziyun Wang$^{1}$\thanks{~~The first two authors contribute equally to this work.} , Xuan Liu$^{2*}$\thanks{~~This work is done during Xuan Liu's internship at Tencent.} , Peiji Yang$^{1}$ , Shixing Liu$^{1}$ , Zhisheng Wang$^{1}$\\
  Tencent, Shenzhen, China \\
  $^{1}$\texttt{\{billzywang, peijiyang, shixingliu, plorywang\}@tencent.com} \\
  $^{2}$\texttt{lxstephenlaw@gmail.com} \\}
\date{}

\begin{document}
\maketitle
\begin{abstract}
Cross-lingual text classification aims at training a classifier on the source language and transferring the knowledge to target languages, which is very useful for low-resource languages. Recent multilingual pretrained language models (mPLM) achieve impressive results in cross-lingual classification tasks, but rarely consider factors beyond semantic similarity, causing performance degradation between some language pairs. In this paper we propose a simple yet effective method to incorporate heterogeneous information within and across languages for cross-lingual text classification using graph convolutional networks (GCN). In particular, we construct a heterogeneous graph by treating documents and words as nodes, and linking nodes with different relations, which include part-of-speech roles, semantic similarity, and document translations. Extensive experiments show that our graph-based method significantly outperforms state-of-the-art models on all tasks, and also achieves consistent performance gain over baselines in low-resource settings where external tools like translators are unavailable.
\end{abstract}

\section{Introduction}

\begin{figure*}[t]
    \centering
    \includegraphics[width=\textwidth]{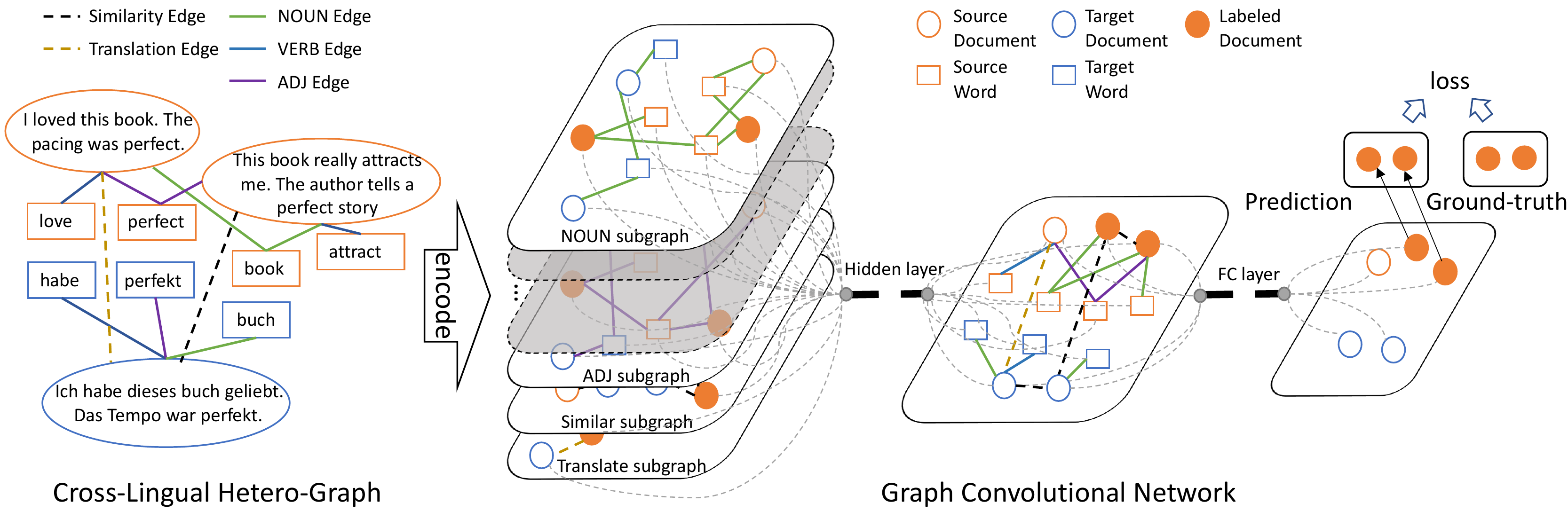}
    \caption{Illustration of our Cross-Lingual Heterogeneous GCN (CLHG) framework. For simplicity, only some POS tags are plotted in this graph. We recommend to view this figure in color as we use different colors to indicate different languages and edge types.}
    \label{fig:framwork}
\end{figure*}

The success of recent deep learning based models on text classification relies on the availability of massive labeled data \citep{conneau-etal-2017,tian-etal-2020,guo-2020}. However, labeled data are usually unavailable for many languages, and hence researchers have developed the setting where a classifier is only trained using a resource-rich language and applied to target languages without annotated data \cite{xu-etal-cross-2016, chen2019transfer, fei-2020}.
The biggest challenge is to bridge the semantic and syntactic gap between languages. Most existing methods explore the semantic similarity among languages, and learn a language-agnostic representation for documents from different languages \cite{chen-2018, zhang-2020}. This includes recent state-of-the-art multilingual pretrained language models (mPLM) \cite{devlin-etal-2019,conneau-2019}, which pretrain transformer-based neural networks on large-scale multilingual corpora. The mPLM methods show superior cross-lingual transfer ability in many tasks \cite{wu-2019}. However, they do not explicitly consider syntactic discrepancy between languages, which may lead to degraded generalization performance on target languages \citep{ahmad-etal-2019,xtreme-2020}.

On the other hand, there usually exists sufficient unlabeled target-language documents that come naturally with rich information about the language and the task. However, only a handful of previous researches have taken advantage of the unlabeled data \cite{wan-2009, dong2019robust}. 

To integrate both semantic and syntactic information within and across languages, we propose a graph-based framework named Cross-Lingual Heterogeneous GCN (CLHG). Following the work of TextGCN \cite{textgcn-2019}, we represent all the documents and words as graph nodes, and add different types of information into the graph. We utilize mPLM to calculate the representation of all the nodes, and connect documents nodes with semantically similar ones to extract the knowledge in mPLM. Words are connected with documents based on the co-occurrences as in previous works. However, we choose to separate different word-doc edges by part-of-speech (POS) tags of words to inject some shallow syntactic information into the graph, as POS taggers are one of the most widely accessible NLP tools, especially for low-resource languages. In-domain unlabeled documents are added to the graph if available. To further absorb in-domain language alignment knowledge, we utilize machine translation to create translated text nodes. The text classification task is then formalized as node classification in the graph and solved with a heterogeneous version of Graph Convolutional Networks \cite{kipf-2017}.
Our contributions are summarized as follows:

(1) We propose a graph-based framework to easily comprise heterogeneous information for cross-lingual text classification, and design multiple types of edges to integrate all these information. To the best of our knowledge, this is the first study to use heterogeneous graph neural networks for cross-lingual classification tasks.
(2) We conduct extensive experiments on 15 tasks from 3 different datasets involving 6 language pairs. Results show that our model consistently outperforms state-of-the-art methods on all tasks without any external tool, and achieves further improvements with the help of part-of-speech tags and translations.

\section{Related Works}

Traditional methods for cross-lingual classification usually translate the texts \cite{wan-2009} or the classifier model \cite{xu-etal-cross-2016} with external aligned resources such as bilingual dictionaries \cite{andrade-2015,shi-2010} or parallel corpora \cite{duh2011machine,zhou-2016,xu-yang-2017}. Recent works focus on learning a shared representation for documents of different languages, including bilingual word embeddings \cite{zou-2013,ziser-2018, chen-2018}, common subword representations \cite{zhang-2020}, and multilingual pretrained language models (mPLM) \citep{devlin-etal-2019,conneau-2019,Clark-2020}.

In the past few years, graph neural networks (GNN) have attracted wide attention, and become increasingly popular in text classification \citep{textgcn-2019,linmei-etal-2019,ding-etal-2020,zhang-etal-2020}. These existing work mainly focus on monolingual text classification, except a recent work \citep{li-etal-2020} using meta-learning and graph neural network for cross-lingual sentiment classification, which nevertheless only uses GNN as a tool for meta-learning.

\section{Method}

In this section, we will introduce our CLHG framework, including how to construct the graph and how to solve cross-lingual text classification using heterogeneous GCN.
In general, we first construct a cross-lingual heterogeneous graph based on the corpus and selected features, and next we encode all the texts with multilingual pre-trained language models, then we pass the encoded nodes to the heterogeneous GCN, each layer of which performs graph convolution on different subgraphs separated by different edge types, and aggregates the information together. Finally, the graph neural network outputs the predictions of doc nodes, which will be compared with groundtruth labels during training.
Figure \ref{fig:framwork} shows the overall structure of the framework.

\subsection{Graph Construction}\label{sec:graph-construction}

Inspired by some previous works on GNN-based text classification \cite{textgcn-2019, linmei-etal-2019}, we construct the graph by representing both documents and words from the corpus in both languages as graph nodes, and augment the corpus by including unlabeled in-domain documents from the target language. To extract more information of language alignments, we further use a publicly available machine translation API\footnote{\url{https://cloud.tencent.com/document/api/551/15619}} to translate the documents in both directions. Then two categories of edges are defined in the graph.

\paragraph{Doc-word Edges.} Like TextGCN \cite{textgcn-2019}, documents and words are connected by their co-occurrences. To inject syntactic information more than just co-occurrences, we add part-of-speech (POS) tags to the edges, since different POS roles have different importance in the classification tasks. Adjectives and adverbs are mostly decisive in sentiment classification, while nouns may play a more significant role in news classification.
Therefore, we use POS taggers to tag each sentence and create different types of edges based on the POS roles of the words in the document, which could help GNN to learn different propagation patterns for each POS role.

\paragraph{Doc-doc Edges.} To add more direct connections between documents, we include two types of document level edges. Firstly, we link each document with similar ones by finding K documents with the largest cosine similarity. The embeddings of the documents are calculated using mPLM. Secondly, we connect nodes created by machine translation with their original texts.


\subsection{Heterogeneous Graph Convolution}

After building the heterogeneous cross-lingual graph, we first encode all the nodes using mPLM by directly inputting the text to the mPLM and taking the hidden states of the first token. The encoded node features are fixed during training. Next we apply heterogeneous graph convolutional network (Hetero-GCN) \cite{linmei-etal-2019} on the graph to calculate higher-order representations of each node with aggregated information.

Heterogeneous GCN applies traditional GCN on different sub-graphs separated by different types of edges and aggregates information to an implicit common space.
\begin{equation}
    H^{(l+1)}=\sigma(\sum_{\tau\in\mathcal{T}}\tilde{A_\tau}\cdot H^{(l)}_\tau\cdot W^{(l)}_\tau)
    \vspace{-0.5em}
\end{equation}
where $\tilde{A_\tau}$ is a submatrix of the symmetric normalized adjacency matrix that only contains edges with type $\tau$, $H^{(l)}_\tau$ is the feature matrix of the neighboring nodes with type $\tau$ of each node, and $W^{(l)}_\tau$ is a trainable parameter. $\sigma(\cdot)$ denotes a non-linear activation function, which we use leaky ReLU. Initially, $H^{(0)}_\tau$ is the node feature calculated by mPLM.

Empirically, we use two graph convolution layers to aggregate information within second-order neighbors. Then a linear transformation is applied to the document nodes to get the predictions.

\begin{table*}[t]
    \centering
    \begin{adjustbox}{max width=\textwidth}
    \begin{tabular}{lllllllllllll} \toprule
         \multirow{2}{*}{Method} & \multicolumn{4}{c}{EN $\to$ DE} & \multicolumn{4}{c}{EN $\to$ FR} &  \multicolumn{4}{c}{EN $\to$ JA}\\ \cmidrule(lr){2-5} \cmidrule(lr){6-9} \cmidrule(lr){10-13} 
                          & books         & dvd        & music         & avg  &   books         & dvd        & music         & avg  & books         & dvd        & music         & avg     \\ \midrule
        CLDFA & 83.95 & 83.14 & 79.02 & 82.04 & 83.37 & 82.56 & 83.31 & 83.08 & 77.36 & 80.52 & 76.46 & 78.11 \\
        MVEC  & 88.41 & 87.32 & 89.97 & 88.61 & 89.08 & 88.28 & 88.50 & 88.62 & 79.15 & 77.15 & 79.70 & 78.67 \\
        mBERT & 84.35 & 82.85 & 93.85 & 83.68 & 84.55 & 85.85 & 83.65 & 84.68 & 73.35 & 74.80 & 76.10 & 74.75 \\
        XLM   & 86.85 & 84.20 & 85.90 & 85.65 & 88.10 & 86.95 & 86.20 & 87.08 & 80.95 & 79.20 & 78.02 & 79.39 \\
        XLM-R & 91.65 & 87.60 & 90.97 & 90.07 & 89.33 & 90.07 & 89.15 & 89.52 & 85.26 & 86.77 & 86.95 & 86.33 \\
        CLHG & \textbf{92.70*} & \textbf{88.60*} & \textbf{91.62*} & \textbf{90.97*} & \textbf{90.67*} & \textbf{91.38*} & \textbf{90.45*} & \textbf{90.83*} & \textbf{87.21*} & \textbf{87.33*} & \textbf{88.08*} & \textbf{87.54*} \\ \bottomrule
    \end{tabular}
    \end{adjustbox}
    \caption{Sentiment classification accuracy (\%) on Amazon Review dataset. * shows the result is significantly better than XLM-R baseline with p-value $\leq0.05$.}
    \label{tab:sentiment-results}
\end{table*}

\begin{table}[t]
    \centering
    \small
    \begin{tabular}{lllll} \toprule
        Method & DE & FR & ES & RU \\ \midrule
        mBERT    & 82.6 & 78.0 & 81.6 & 79.0 \\
        XLM-R    & 84.5 & 78.2 & 83.2 & 79.4 \\
        Unicoder & 84.2 & 78.5 & 83.5 & 79.7 \\
        XLM-R (ours)    & 83.99 & 78.66 & 83.27 & 80.42 \\
        CLHG & \textbf{85.00}$^+$ & \textbf{79.58}* & \textbf{84.80}* & \textbf{80.91}$^+$ \\ \bottomrule
    \end{tabular}
    \caption{Classification accuracy (\%) on XGLUE News Classification. We re-run the XLM-R model and also report our reproduced results. * shows the result is significantly better than XLM-R baseline with p-value $\leq0.05$, and $^+$ indicates p-value $\leq0.1$.}
    \label{tab:xglue-nc-results}
\end{table}

\begin{table}[t]
    \centering
    \small
    \begin{tabular}{lll} \toprule
        Method & EN $\to$ ES & EN $\to$ TH \\ \midrule
        CoSDA\_ML+mBERT & 94.80 & 76.80 \\
        CoSDA\_ML+XLM & 90.30 & 86.70 \\
        mBERT & 74.91 & 42.97 \\
        XLM & 62.30 & 31.60 \\
        XLM-R & 94.38 & 85.17 \\
        CLHG & \textbf{96.81}* & \textbf{89.71}* \\ \bottomrule
    \end{tabular}
    \caption{Intent classification accuracy (\%) on multilingual SLU dataset. * shows the result is significantly better than XLM-R baseline with p-value $\leq0.05$.}
    \label{tab:intent-results}
\end{table}

\section{Experiments}

We evaluate our framework on three different classification tasks, including Amazon Review sentiment classification \cite{prettenhofer-2010}, news category classification from XGLUE \cite{Liang2020XGLUEAN}, and intent classification on a multilingual spoken language understanding (SLU) dataset \cite{schuster2019cross}. More details of each dataset is provided in the appendix. For all the tasks, we use only the English samples for training and evaluate on other 6 languages, which are German (DE), French (FR), Russian (RU), Spanish (ES), Japanese (JA), and Thai (TH).

\subsection{Experiment Setting}
In all our experiments, we use two-layer GCN with hidden size $512$ and output size $768$. Each document is connected with 3 most similar documents. The model is trained using the AdamW optimizer \cite{loshchilov2019decoupled} with a learning rate of $2\times10^{-5}$ and batch size $256$. We train the GCN for at most $15$ epochs and evaluate the model with best performance on validation set. XLM-RoBERTa \cite{conneau2020unsupervised} is used to encode all the documents and words, which is finetuned on the English training set of each task for $2$ epochs with batch size $32$ and learning rate $4\times10^{-5}$.
We set the max length as $128$ for intent classification, and $512$ for the other two tasks. Each experiment is repeated $3$ times and the average accuracy is reported. All the experiments are conducted on an NVIDIA V100 GPU \footnote{Our codes are available at \url{https://github.com/TencentGameMate/gnn_cross_lingual}.}. 

For part-of-speech tagging, we adopt different taggers for each language \footnote{We choose Stanford POS Tagger \cite{toutanova2003feature} for EN, Spacy (\url{https://spacy.io}) for DE, FR and ES, MeCab (\url{https://taku910.github.io/mecab/}) for JA, tltk (\url{https://pypi.org/project/tltk/}) for TH, and nltk (\url{https://www.nltk.org}) for RU.} and map all the tags to Universal Dependency (UD) tagset \footnote{\url{https://universaldependencies.org}}.

\subsection{Baselines}
Our method is compared with different multilingual pretrained models finetuned for each task, which include multilingual BERT \cite{devlin-etal-2019}, XLM \cite{conneau-2019} and XLM-RoBERTa \cite{conneau2020unsupervised}, and also with published state-of-the-art results on each dataset.

\paragraph{Amazon Review.} CLDFA \cite{xu-yang-2017} utilizes model distillation and parallel corpus to transform a model from source to target language. MVEC \cite{fei-2020} refines the shared latent space of mPLM with unsupervised machine translation and a language discriminator.

\paragraph{XGLUE News Classification.} Unicoder \cite{huang2019unicoder} is another mPLM proposed recently, and is used as the baseline method provided alongside XGLUE benchmark.

\paragraph{Multilingual SLU.} CoSDA-ML \cite{qin2020cosda} is a data augmentation framework that automatically generates code-switching documents using a bilingual dictionary, which is used when finetuning language models on downstream tasks.

\begin{table*}[t]
    \centering
    \small
    \begin{tabular}{lccccccccc} \toprule
        & XLM-R & 1 & 2 & 3 & 4 & 5 & 6 & 7 & full model \\ \midrule
        word-doc & & \checkmark & & \checkmark & \checkmark & \checkmark & \checkmark & \checkmark & \checkmark \\
        POS tags & & & & & & \checkmark & \checkmark & \checkmark & \checkmark \\
        translation edges & & & \checkmark & & \checkmark & & \checkmark & \checkmark & \checkmark \\
        similarity edges & & & \checkmark & \checkmark & \checkmark & \checkmark & & \checkmark & \checkmark \\
        unlabeled & & & \checkmark & \checkmark & \checkmark & \checkmark & \checkmark & & \checkmark
          \\ \midrule
        EN $\to$ DE & 90.01 & 87.60 & 90.60 & 90.75 & 90.77 & 90.67 & 89.90 & \textbf{91.26} & 90.97 \\
        EN $\to$ FR & 89.52 & 90.62 & 89.95 & 90.65 & 90.70 & 90.37 & 89.82 & 90.85 & \textbf{90.92} \\
        EN $\to$ JA & 86.61 & 86.26 & 87.19 & 87.31 & 87.35 & 87.44 & 87.18 & 86.57 & \textbf{87.54} \\
         \bottomrule
    \end{tabular}
    \caption{Ablation study results. The left-most column shows the results of finetuned XLM-R, and others each indicates a variant in graph construction. We conduct experiments on the Amazon Review dataset and report the average accuracy across three domains.}
    \label{tab:ablation-results}
\end{table*}

\subsection{Results and Analysis}
The results are provided in table \ref{tab:sentiment-results} \ref{tab:xglue-nc-results} and \ref{tab:intent-results} for each dataset. Our method significantly outperforms state-of-the-art baselines and achieves consistent improvements over XLM-R model. The most performance gain is achieved on the multilingual SLU dataset. Different from the other two, this dataset consists of short texts, and thus the created graph is much cleaner and more suitable for GCN to model.

To verify that the improvement does not come barely from the external data, we conduct another experiment that adds the translated data to the training set and finetunes the baseline XLM-R model on Amazon Review dataset. The results showed very slight improvement (0.09\% on average), showing that XLM-R cannot directly benefit from external training data.

Ablation studies are performed on Amazon Review dataset to analyze the effectiveness of different graph structures.
From the results provided in table \ref{tab:ablation-results}, variant 1 containing a homogeneous graph with only word-doc edges (same as TextGCN) performs the worst, while adding more information leads to better performance in general.
Comparing variants 4-7, similarity demonstrates to be the most important among all added information. Similarity edges help the model to converge faster and learn better as well, since they provide a ``shortcut'' between documents that is highly likely to be in the same category. 
Variant 7 shows that unlabeled corpus play an important role in EN$\to$JA setting, but less effective when transferring between similar languages, since unlabeled data inevitably contain some noise and do not provide much help for linguistically-closer languages. Variant 4 also shows that POS tags are more helpful for distant language pairs like EN$\to$JA, and our added experiment on EN$\to$TH shows greater impact of POS tags (89.71$\to$88.06 when removing POS tags).
Additionally, we test a variant without any external tool that requires training resources in the target language. Variant 3 does not rely on POS tagger or translation service, and still outperforms the XLM-R baseline with a large margin. This demonstrates that our method can be adopted for real low-resource languages without good tagger or translator.



\section{Conclusion}

In this study, we propose a novel graph-based method termed CLHG to capture various kinds of information within and across languages for cross-lingual text classification.
Extensive experiments illustrate that our framework effectively extracts and integrates heterogeneous information among multilingual corpus, and these heterogeneous relations can enhance existing models and are instrumental in cross-lingual tasks. There may exist some better semantic or syntactic features and combinations of features, which we leave as a future work to explore. We also wish to extend our GNN-based framework to different NLP tasks requiring knowledge transfer and adaptation in the future.


\section*{Acknowledgement}
We thank all the anonymous reviewers for their careful reading and helpful comments. We thank our colleagues for their thoughtful and inspiring discussions during the development of this work. We appreciate Tencent Cloud for providing computational resources and technical support.

\bibliographystyle{acl_natbib}
\bibliography{acl2021}

\clearpage
\appendix

\begin{table*}[t]
    \centering
    \small
    \begin{tabular}{lclrrrrr}
    \toprule
    dataset & \#category & language & \#train & \#valid & \#test & \#unlabeled & avg. length \\ \midrule
    \multirow{4}{*}{Amazon Review (Books)} & \multirow{4}{*}{2} & English & 2,000 & / & 2,000 & 50,000 & 168.31 \\
     &  & German & 2,000 & / & 2,000 & 165,457 & 151.27 \\
     &  & French & 2,000 & / &2,000 & 32,868 & 123.84 \\
     &  & Japanese & 2,000 & / &2,000 & 169,756 & 155.05 \\ \midrule
     \multirow{4}{*}{Amazon Review (Dvd)} & \multirow{4}{*}{2} & English & 2,000  & / & 2,000 & 30,000 & 167.31 \\
     &  & German & 2,000 & / &2,000 & 91,506 & 158.58 \\
     &  & French & 2,000 & / &2,000 & 9,356 & 138.89 \\
     &  & Japanese & 2,000 & / &2,000 & 68,324 & 150.87 \\ \midrule
     \multirow{4}{*}{Amazon Review (Music)} & \multirow{4}{*}{2} & English & 2,000 & / &2,000 & 25,220 & 146.18 \\
     &  & German & 2,000 & / &2,000 & 60,382 & 143.50 \\
     &  & French & 2,000 & / &2,000 & 15,940 & 142.21 \\
     &  & Japanese & 2,000 & / &2,000 & 55,887 & 131.62 \\ \midrule
     \multirow{5}{*}{XGLUE NC} & \multirow{5}{*}{10} & English & 100,000 & 10,000 & 10,000 & / & 553.65 \\
     &  & German & / & 10,000 & 10,000 & / & 484.69 \\
     &  & French & / & 10,000 & 10,000 & / & 567.86 \\
     &  & Spanish & / & 10,000 & 10,000 & / & 533.09 \\
     &  & Russian & / & 10,000 & 10,000 & / & 426.80 \\ \midrule
     \multirow{3}{*}{Multilingual SLU} & \multirow{3}{*}{12} & English & 30,521 & 4,181 & 8,621 & / & 8.05 \\
     &  & Spanish & 3,617 & 1,983 & 3,043 & / & 8.74 \\
     &  & Thai & 2,156 & 1,235 & 1,692 & / & 5.12 \\
     \bottomrule
    \end{tabular}
    \caption{Summary statistics of the datasets. The average length shows the average number of words in each document.}
    \label{tab:dataset}
\end{table*}

\section{Datasets}
Here we introduce the three datasets we use in our experiments. A summary of the statistics for three datasets are provided in table \ref{tab:dataset}.

\paragraph{Amazon Review \footnote{\url{https://webis.de/data/webis-cls-10.html}}} This is a multilingual sentiment classification dataset covering 4 languages (English, German, French, Japanese) on three domains (\textit{Books, DVD}, and \textit{Music}). The original dataset contains ratings of 5-point scale. Following previous works \cite{xu-yang-2017, fei-2020}, we convert the ratings to binary labels with threshold at 3 points.
Since English is not used for testing, we follow the previous works and re-construct the training and validation set by combining the English training and test set. The new training set contains 3,200 randomly sampled documents. We use the training set of target languages for validation. This dataset also provides large amount of unlabeled data for each language, which is used in our framework.

\paragraph{XGLUE News Classification \footnote{\url{https://microsoft.github.io/XGLUE/}}} This is a subtask of a recently released cross-lingual benchmark named XGLUE. This subtask aims at classifying the category of a news article, which covers 10 categories in 5 languages, including English, Spanish, French, German and Russian. This dataset does not contain unlabeled data for target languages, so we do not add unlabeled documents in our graph either.

\paragraph{Multilingual SLU \footnote{\url{https://fb.me/ multilingual_task_oriented_data}}} This dataset contains short task-oriented utterances in English, Spanish and Thai across weather, alarm and reminder domains. We evaluate our framework on the intent classification subtask, which has 12 intent types in total. 
For each target language, we use the original training set as unlabeled data added in the graph.

\section{Hyperparameter Search}
We perform a grid search to pick the best combination of hyperparameters. The hidden size and output size are chosen among \{384, 512, 768\}, and the learning rate within \{$1\times10^{-5}$, $2\times10^{-5}$, $5\times10^{-5}$, $1\times10^{-4}$\}. For the XLM-R baseline, we also tune the learning rate within \{$2\times10^{-5}$, $4\times10^{-5}$, $5\times10^{-5}$\} and number of epochs from 2 to 5. Among all the combination of hyperparameters, we pick the values with the best performance on the German training set from \textit{Books} domain of Amazon Review dataset, and use the same set of hyperparameters for all our experiments. The maximum length of XLM-R model is chosen based on statistics of the datasets.

\end{document}